\documentclass{article}

\usepackage[final]{neurips_2019}

\usepackage[utf8]{inputenc}
\usepackage[T1]{fontenc}
\usepackage{hyperref}
\usepackage{url}
\usepackage{booktabs}
\usepackage{amsfonts}
\usepackage{nicefrac}
\usepackage{microtype}
\usepackage{graphicx}
\usepackage{xcolor}
\usepackage{lipsum}
\usepackage{multirow}
\usepackage{enumitem}
\usepackage{wrapfig}
\usepackage{caption}

\usepackage{tcolorbox}
\newtcolorbox{customtb}{
width=1.\linewidth,
boxrule=.5pt,
colframe=black
}

\title{
  Temporal Grounding of Activities using \\
  Multimodal Large Language Models
}

\author{
  Young Chol Song \\
  Stanford University \\
  \texttt{youngcs@stanford.edu} \\
}

\begin{document}

\maketitle

\begin{abstract}
Temporal grounding of activities, the identification of specific time intervals of actions within a larger event context, is a critical task in video understanding. Recent advancements in multimodal large language models (LLMs) offer new opportunities for enhancing temporal reasoning capabilities. In this paper, we evaluate the effectiveness of combining image-based and text-based large language models (LLMs) in a two-stage approach for temporal activity localization. We demonstrate that our method outperforms existing video-based LLMs. Furthermore, we explore the impact of instruction-tuning on a smaller multimodal LLM, showing that refining its ability to process action queries leads to more expressive and informative outputs, thereby enhancing its performance in identifying specific time intervals of activities. Our experimental results on the Charades-STA dataset highlight the potential of this approach in advancing the field of temporal activity localization and video understanding.
\end{abstract}

\section{Introduction} \label{introduction}
We address the problem of temporal grounding of activities, also known as temporal activity localization~\citep{gao2017tall} or video moment retrieval~\citep{cui2022video}. This task involves determining the interval of a specific target activity within a longer context of activities, where the context is provided as a video and the target activity is described in text. However, accurately grounding activities in time poses significant challenges due to the complex and dynamic nature of real-world events, variability in human behavior, and the need for fine-grained understanding of temporal relationships.

Recent advancements in multimodal large language models (LLMs)~\citep{liu2023llava, zhang2024llamaadapter} have opened new avenues for addressing this task. Many methods for temporal grounding primarily rely on specialized architectures and task-specific training~\citep{zhang2020learning, cui2022video}, which can limit their generalizability and scalability. In contrast, multimodal LLMs, with their broad knowledge base and reasoning capabilities~\citep{huang2023vtimellm, li2024groundinggptlanguage}, have the potential to overcome some of these limitations.

In this paper, we explore the use of pretrained and tuned multimodal LLMs for temporal grounding of fine-grained activities, such as `putting on shoes' or `pouring a drink.' We demonstrate that a combination of image-based and text-based LLMs achieves higher accuracy in temporal activity localization than a purely video-based LLM of a similar parameter size. Additionally, we show that instruction-tuning a smaller multimodal LLM on a dataset generated by GPT-4 Vision~\citep{openai2023gpt4} not only qualitatively enhances the model's visual description of activities but also improves its performance in accurately identifying the time intervals of these activities.

\begin{figure}[t]
\centering
\includegraphics[width=1.\textwidth]{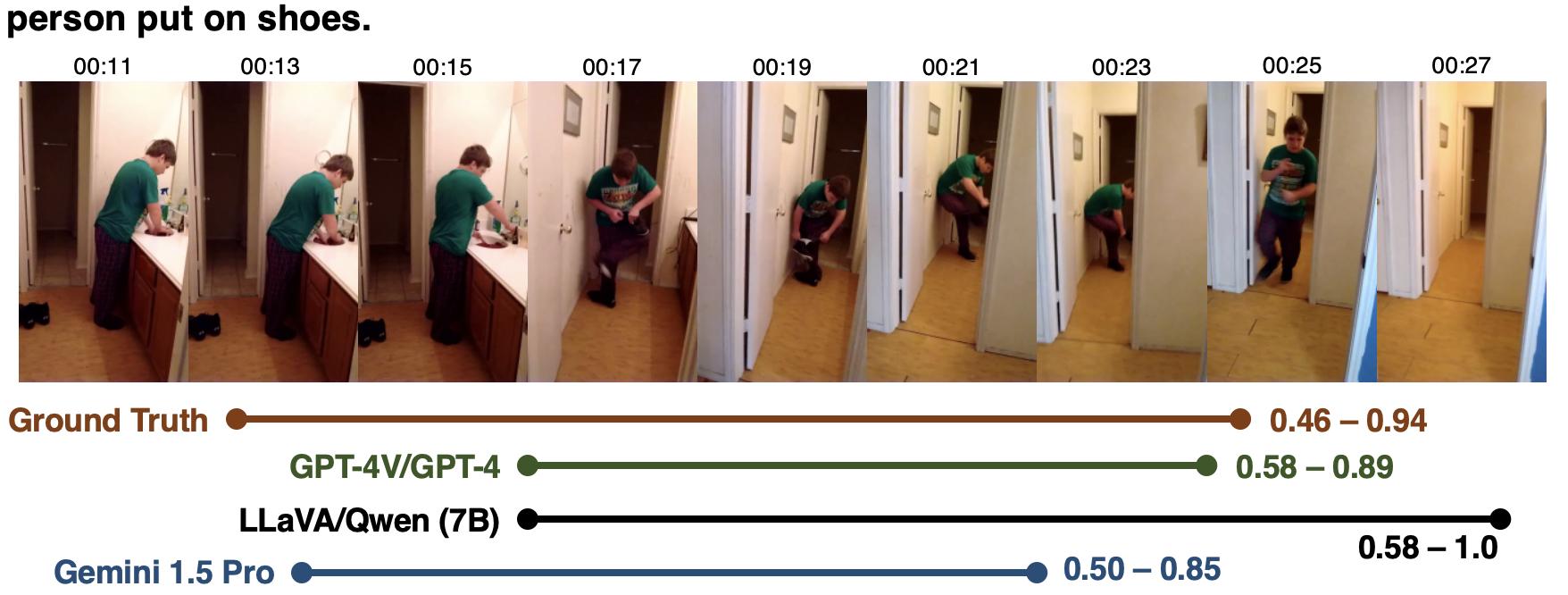}
\captionsetup{justification=centering}
\caption{Comparing different model outputs of temporal activity localization\\
Numbers to the right are normalized intervals of the activity ``person put on shoes.''}
\label{activity_illustration}
\end{figure}

\section{Related Work} \label{relatedwork}

\paragraph{Multimodal Large Language Models} Historically, multimodal vision and language systems for activity or action recognition focused primarily on solving challenges in vision. While some models emphasized aspects of language modeling~\citep{li2017learning, wang2020temporally, radford2021learning}, the role of language was often reduced to providing simple captions or labels in many cases. With the advancement of pretrained large language models (LLMs), models have emerged that fully utilize the strength of LLMs~\citep{li2023blip, alayrac2022flamingo}, and their ability to query, interact, and reason using text within the context of the image~\citep{liu2023improved, liu2023llava, chen2023minigptv2, chen2023llava}. In the past year, models that can directly take videos as input, such as Video-ChatGPT~\citep{maaz2023video}, have started to emerge~\citep{zhang2023video, li2023videochat, reid2024gemini}, enabling pretrained models that can conduct temporal grounding and reasoning out of the box without any customization or tuning.

\paragraph{Temporal Activity Localization} Early approaches to temporal activity localization focused on computer vision approaches using convolutional neural networks and sentence embeddings~\citep{gao2017tall} or LSTMs~\citep{anne2017localizing, zhang2020learning} on the language side. In recent years, methods using attention mechanisms, such as the Transformer model have started to emerge~\citep{yuan2019semantic, cui2022video}, offering improved performance. With advances in video-based LLMs, models specifically tuned for temporal grounding~\citep{huang2023vtimellm, qian2024momentor}, and more generally tuned for both spatial and temporal grounding, have started gaining traction~\citep{li2024groundinggptlanguage}.

\section{Approach} \label{approach}
We introduce a two-stage approach for temporal activity grounding. The initial stage utilizes the image-based multimodal LLM to generate detailed descriptions of actions occurring in the individual video frames. In the second stage, we feed the responses from the multimodal LLM to a text-based LLM along with the activity query to pinpoint when the activity occurs within the sequence.

\subsection{Multimodal LLM (Image-based)} \label{mmllm}
\begin{figure}[t]
\centering
\includegraphics[width=1.\textwidth]{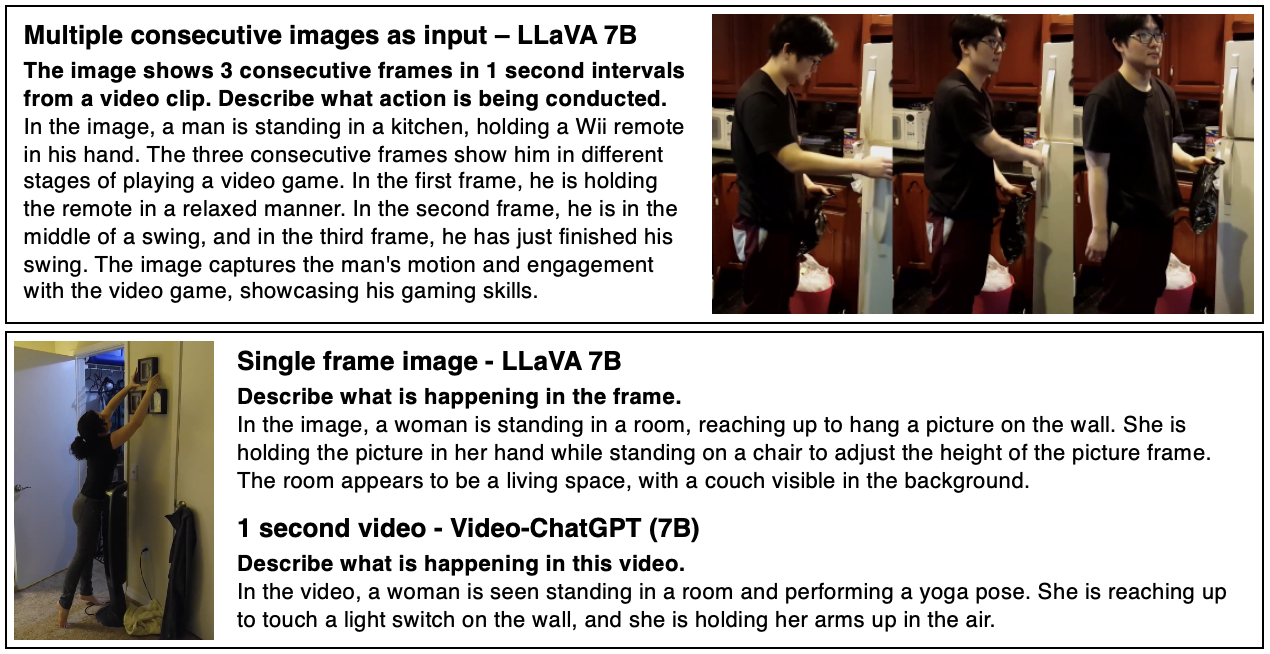}
\captionsetup{justification=centering}
\caption{Responses from a multimodal LLM based on the type of input: \\
consecutive frames (top), single frame (mid), short video (bottom)}
\label{different_types}
\end{figure}

The first stage consists of a multimodal LLM capable of image input, such as LLaMA-Adapter~\citep{zhang2024llamaadapter}, LLaVA~\citep{liu2023llava}, or GPT-4 Vision~\citep{openai2023gpt4}. The role of this stage is to describe the activities occurring in each individual video frame. In our current approach, we use single frames from a video as input as it provides the best quality response. We have experimented with providing different types of inputs to the multimodal LLM and discuss this further in Section \ref{analysis}. Examples of such outputs are shown in Figure \ref{different_types}. We use two different strategies when providing prompts to the multimodal LLM, which we will refer to as the \textit{activity prompt} and the \textit{general prompt}.

\paragraph{Activity prompt} This type of prompt provides the context of the action during the prompting of the multimodal LLM. This enables the model to search for the particular action in the frame(s) and provide information with regards to the particular action in question. However, the activity prompt comes with a complication where, if there are multiple activities to identify in a particular video sequence, separate prompts will be required for each frame image/activity combination.

\begin{customtb}
\texttt{[Image frame] This is one frame from a longer video clip. The video clip includes the action described as [activity query] however, the frame may or may not include this action. Briefly explain what action or actions the person is conducting in the frame.}
\end{customtb}
{\centering Activity prompt\par}

\paragraph{General prompt} The second way to prompt the multimodal LLM is to provide the model with a generic prompt asking it to describe the scene. This saves on having to query multiple times for each frame image depending on the number of activities in the scene and may be useful when frames need to be processed beforehand. However, there is a tradeoff in accuracy as information about the particular activity in question may not be included in the response.
\begin{customtb}
\texttt{[Image frame] Describe what is happening in the frame.}
\end{customtb}
{\centering General prompt\par}

For our instruction-tuning experiments, we will evaluate models using the \textit{general prompt} strategy to best illustrate how instruction-tuning a relatively smaller model (LLaVA 7B) with instruction-tuning data generated from a larger model (GPT-4 Vision) can provide better results due to enhanced descriptive outputs. For our model comparison experiments, we will use the \textit{action prompt} strategy to prioritize model performance. In addition, we use the STA-Subset dataset (described in Section~\ref{data}) for the model comparison experiments, as the image/activity combinations required for running these experiments using \textit{action prompts} are not as prohibitive as the former.

\subsection{Text-based LLM}
The second stage uses a text-based LLM, which utilizes descriptions aggregated with each frame output from the first stage, and executes the task of predicting the most likely interval where the particular action would happen. The target action in question is necessarily included in this prompt.

\begin{customtb}
\texttt{The following are descriptions of actions for frames extracted 1 second apart from a video clip:\\
 $\ast$ Frame 1: [Description from frame 1]\\
 $\ast$ Frame 2: [Description from frame 2]\\
 $\ast$ $\cdots$\\
The action [activity query] has occurred in the video clip. What interval is the action most likely to start and end? Provide your best guess by providing the start and end frame numbers in json format.}
\end{customtb}
{\centering Template of a prompt used to query the text-based LLM\par}

We explicitly request the responses in JSON format for parsing convenience; however, the text-based LLM is not bound to this convention and may generate non-compliant JSON responses. We employ a heuristic-based parser, which uses criteria such as the presence of specific keywords or patterns, to extract the start and end frames from the entire response text if the frame numbers from the JSON cannot be extracted.

In our experiments, we compare different text-based LLMs for our second stage processing, such as OpenAI's GPT-4~\citep{openai2023gpt4}, Qwen 14B, and Qwen 7B~\citep{qwen}. As the aggregated descriptions from multiple video frames became quite lengthy, we were limited to models that could handle context lengths that covered text from our longest video. Models with context lengths of 4096 tokens or less generally had trouble processing our descriptions, which discouraged our use of LLaMA or LLaMA-based models such as Vicuna.

\subsection{Video-based LLM}
For comparison purposes, we also use Gemini 1.5 Pro, a multimodal LLM that can take videos directly as input. We use a prompt similar to the one used for the text-only LLM but without providing the image description context.

\begin{customtb}
\texttt{The action [activity query] has occurred in the video clip. In what interval is the action most likely to occur? Please provide the numbers for the start and end timestamps.}
\end{customtb}
{\centering Template of a prompt used to query the video-based LLM\par}

\subsection{Constraints}
A major limiting factor in our approach was the amount of compute resources and cost needed to process \textasciitilde100k images via the multimodal LLM for each experiment. To mitigate this, while we use the Charades-STA full dataset for our instruction-tuning experiments on our 7B models, we employ the \textit{general prompting} strategy as described above. For our model comparison experiments, we use the \textit{activity prompting} strategy to focus on performance but select a smaller test set (STA-Subset, described in Section \ref{data}) during evaluations.

\section{Experiments} \label{experiments}
We conduct experiments exploring the different aspects of temporal activity localization using multimodal LLMs. To demonstrate the benefits of instruction-tuning in multimodal LLMs, we create an instruction-tuning dataset from a larger multimodal LLM (e.g., GPT-4 Vision) and instruction-tune a smaller multimodal LLM (e.g., LLaVA 7B) based on this dataset to enhance its descriptive power in responding to activity-related prompts.

We also formally evaluate our two-stage approach to temporal activity localization using pretrained and instruction-tuned LLMs, where the first stage involves a multimodal LLM tasked with describing activities within frames from a video, and the second stage employing a language-only LLM for reasoning about the timing of the activity within the video clip.

We compare the performance metrics of our tuned model against similar parameter-sized video-based LLMs and evaluate our results by comparing the base multimodal LLM to our tuned model, showing our model performs better in the task of grounding activities temporally via our two-stage LLM approach. We also use LLMs of various types and sizes to compare our two-stage approach, showing how the multimodal and text-based stages perform differently when the underlying LLMs change.

\subsection{Data} \label{data}

\paragraph{Instruction-tuning Dataset} A set of \textasciitilde19,000 activity-related queries generated by GPT-4 Vision using frames randomly extracted from the training set portion of the Charades-STA dataset~\citep{gao2017tall}. These queries are designed to help the model become more descriptive in responding to activity-related prompts. Each sample is split into two parts, the first prompt always asks for a description of what is happening in the frame, followed by an optional Q\&A sequence asking for specific activity-related questions about the frame. The samples were created by using GPT-4 Vision to analyze the frame and provide a verbose description, and generate questions and their answers regarding activities occurring in the frame.

\paragraph{Charades-STA Activity Dataset} We use an activity dataset with longer duration activities for our evaluations, which were also similarly used in~\citet{cui2022video}. The Charades~\citep{sigurdsson2016hollywood} dataset contains 9848 videos of 157 action classes in total, with 27847 textual descriptions. The Charades-STA~\citep{gao2017tall} dataset (abbreviated as \textit{STA} in subsequent sections) is a selected portion of Charades that contains sentence temporal annotations, with a total of 6672 videos of 16128 sentence temporal annotations, where 3720 annotations from 1334 videos are used for testing. For image inputs, we extracted images from each video at a rate of 1 frame per second (fps), resulting in an average of 31 frames per video. We use the Charades-STA test set for our instruction-tuning evaluations for the purpose of comparing our results with existing video-based LLM evaluations.

\paragraph{Charades-STA Subset (STA-Subset)} We defined a smaller subset of the Charades-STA activity dataset, denoted as \textit{STA-Subset} in the subsequent sections, consisting of 128 videos each with one sentence temporal annotation. This results in a total of 4000 frames when extracted into images at 1 fps. The smaller subset is used due to the prohibitive cost in both compute and time in running multimodal LLMs over each frame/sentence temporal annotation combination in the original Charades-STA dataset. We use this subset for comparing between different LLMs in our two-stage approach comparison evaluations.

\subsection{Evaluation method}
We use the metric recall of threshold-bounded temporal intersection over union, denoted as ``R@$n$, IoU=$m$''~\citep{gao2017tall}. In all of our quantitative evaluations, we use $n=1$, and abbreviate the metric as R@IoU=$m$. The variable $m$ in R@IoU can have fractional values, indicating the threshold of overlap of the temporal prediction of the action with the ground truth. As used in~\citet{gao2017tall}, evaluations are conducted using R@IoU thresholds of $\{0.3, 0.5, 0.7\}$, and a single value averaging the different R@IoU values is noted as mean averaged IoU (mIoU).

\subsection{Experimental details}

\paragraph{Instruction-tuning} For the baseline, we ran inference on a non-modified LLaVA 7B (\texttt{LLaVA-1.5-7B}) model using the \textit{general prompt} strategy introduced in Section \ref{mmllm}. We then instruction-tuned the model using the dataset described in Section \ref{data} using the \texttt{finetune\_task\_lora.sh} script\footnote{\url{https://github.com/haotian-liu/LLaVA/blob/main/scripts/v1_5/}} provided by LLaVA over $1$ epoch, with a learning rate of $2e^{-4}$, for a total of $1.4$k steps. Total time for instruction-tuning took around $2$ hours on an NVIDIA H100 GPU. Due to the longer context generated by describing each frame in the video, we chose Qwen 7B~\citep{qwen} for the text-based LLM as it has a context length of 8192 tokens. Using LLaMA 2~\citep{touvron2023llama}, we ran into issues with context length, as the model only has half the context length of that of Qwen models.

\paragraph{Two-stage LLM runs} Next, we compare metrics of different combinations of multimodal LLM and text-based LLMs as listed in Table \ref{table:perfmetrics}. We use the \textit{activity prompting} strategy in our multimodal LLMs to optimize for performance. For comparison, we run models on the state-of-the-art task-specific model (ViGA) by~\citet{cui2022video} and a state-of-the-art video-based LLM~\citep{reid2024gemini}, Gemini.

\begin{table}[t]
\begin{center}

\begin{tabular}{ p{1.7cm}|p{2cm}|p{.7cm}|p{.7cm}|p{.7cm}|p{.7cm}|p{.7cm}|p{.7cm}|p{.7cm}|p{.7cm} }
 \hline
 \multicolumn{1}{c}{Multimodal} & \multicolumn{1}{c||}{Text-based} & \multicolumn{4}{c|}{STA ($n$=3720) R@IoU} & \multicolumn{4}{c}{STA-Subset ($n$=128) R@IoU} \\
 \multicolumn{1}{c}{LLM} & \multicolumn{1}{c||}{LLM} & \multicolumn{1}{c}{0.3} & \multicolumn{1}{c}{0.5} & \multicolumn{1}{c}{0.7} & \multicolumn{1}{c|}{mIoU} & \multicolumn{1}{c}{0.3} & \multicolumn{1}{c}{0.5} & \multicolumn{1}{c}{0.7} & \multicolumn{1}{c}{mIoU} \\
 \hline
 \multicolumn{2}{c||}{VideoLLaMA (7B)} & \centering 10.4 & \centering 3.8 & \centering 0.9 & \multicolumn{1}{c|}{7.1} & \multicolumn{1}{c}{} & \multicolumn{1}{c}{} & \multicolumn{1}{c}{} & \multicolumn{1}{c}{} \\
 \multicolumn{2}{c||}{VideoChat (7B)} & \centering 9.0 & \centering 3.3 & \centering 1.3 & \multicolumn{1}{c|}{6.5} & \multicolumn{1}{c}{} & \multicolumn{1}{c}{} & \multicolumn{1}{c}{} & \multicolumn{1}{c}{} \\
 \multicolumn{2}{c||}{VideoChatGPT (7B)} & \centering 20.0 & \centering 7.7 & \centering 1.7 & \multicolumn{1}{c|}{13.7} & \multicolumn{1}{c}{} & \multicolumn{1}{c}{} & \multicolumn{1}{c}{} & \multicolumn{1}{c}{} \\
 \multicolumn{1}{c}{LLaVA 7B} & \multicolumn{1}{c||}{Qwen 7B} & \centering 14.1 & \centering 5.6 & \centering 1.8 & \multicolumn{1}{c|}{12.0} & \centering 7.0 & \centering 3.1 & \centering 0.0 & \multicolumn{1}{c}{8.6} \\
 \multicolumn{1}{c}{\textbf{Tuned 7B}} & \multicolumn{1}{c||}{\textbf{Qwen 7B}} & \centering \textbf{20.1} & \centering \textbf{7.9} & \centering \textbf{2.5} & \multicolumn{1}{c|}{\textbf{14.9}} & \centering \textbf{15.6} & \centering \textbf{7.0} & \centering \textbf{0.8} & \multicolumn{1}{c}{\textbf{12.7}} \\
 \multicolumn{1}{c}{GPT-4 Vision} & \multicolumn{1}{c||}{GPT-4} & \multicolumn{1}{c}{} & \multicolumn{1}{c}{} & \multicolumn{1}{c}{} & \multicolumn{1}{c|}{} & \centering 28.9 & \centering 10.9 & \centering 6.3 & \multicolumn{1}{c}{20.7} \\
 \hline
\end{tabular}
\vspace{.5em}
\captionsetup{justification=centering}
\caption{R@IoU for base LLaVA, instruction-tuned LLaVA using the \textit{general prompting} strategy.\\
For comparison, we also show video-based LLM~\citep{huang2023vtimellm}, GPT-4 Vision metrics.}
\label{table:tunemetrics}
\end{center}
\end{table}

For inference in all experiments, LLaVA 7B (\texttt{LLaVA-1.5-7B}\footnote{\url{https://huggingface.co/liuhaotian/llava-v1.5-7b}}) and 16B (\texttt{LLaVA-1.6-Vicuna-13B}\footnote{\url{https://huggingface.co/liuhaotian/llava-v1.6-vicuna-13b}}) models were run on NVIDIA RTX A4500 and A6000 GPUs. Interactions with Qwen 14B (\texttt{Qwen/Qwen1.5-14B-Chat}), Qwen 7B (\texttt{Qwen/Qwen1.5-7B-Chat}) and Gemma 7B (\texttt{google/gemma-7b-it}) models were conducted via the Together AI API\footnote{\url{https://api.together.xyz/}}. GPT-4 Vision (\texttt{gpt-4-1106-vision-preview}) and GPT-4 (\texttt{gpt-4-0125-preview}) models were accessed through the OpenAI API\footnote{\url{https://openai.com/blog/openai-api}}. For GPT-4 Vision, we use low-res mode which resizes the images to 512px $\times$ 512px. LLaVA models use \texttt{CLIP-ViT-L-336px}~\citep{radford2021learning} for its vision encoders, which resize images to 336px $\times$ 336px. We used pretrained ViGA models provided by the authors\footnote{\url{https://github.com/r-cui/ViGA/tree/master}} to generate metrics for ViGA using VGG and I3D features. Access to Gemini 1.5 Pro was available via Google AI Studio’s web interface\footnote{\url{https://aistudio.google.com/}}.

\subsection{Results}

\paragraph{Instruction-tuning metrics} In Table \ref{table:tunemetrics}, we report the R@IoU metrics for the base model and the instruction-tuned model for both the full Charades-STA dataset (STA) and the STA-Subset. For reference, we also include R@IoU metrics for video-based LLMs as described in~\citet{huang2023vtimellm, li2024groundinggptlanguage}, for VideoLLaMA~\citep{zhang2023video}, VideoChat~\citep{li2023videochat} and VideoChatGPT~\citep{maaz2023video}. We observe that the instruction-tuned 7B model performs significantly better than the original LLaVA model and outperforms other video-based 7B LLM models. The instruction-tuned model is more action-focused in its descriptions and is more verbose, enabling the text-based LLM to work with more context compared with the base model. We cover qualitative analysis of instruction-tuning results with examples in Section \ref{analysis}.

\paragraph{Two-stage LLM comparisons} Metrics for each combination of multimodal and text-based LLMs are shown in Table \ref{table:perfmetrics}. Overall, the two-stage GPT-4 Vision with GPT-4 model performed best, mostly outperforming the current state-of-the-art video-based LLM, Gemini 1.5 Pro, except when R@IoU=$0.7$. An example of two-stage LLM output comparisons is illustrated in in Figure \ref{activity_illustration}. As seen in Table \ref{table:perfmetrics}, both the multimodal LLM and text-based LLM contribute in different ways to the temporal activity localization accuracy. There are two interesting observations: in the multimodal LLM, LLaVA 7B seems to perform better than 16B. Also, there appear to be pairs of multimodal LLM and text-based LLMs that work well together and others that don't. For example, given GPT-4 Vision's output, Gemma 7B has a hard time providing proper time interval outputs, many of the outputs in this case are undefined, which significantly affects the resulting metric.

\begin{table}[t] 
\begin{center}

\begin{tabular}{ p{2.5cm} p{3cm}|p{1.2cm}|p{1.2cm}|p{1.2cm}|p{1.2cm} }
 \hline
 \multicolumn{1}{c}{Multimodal} & \multicolumn{1}{c|}{Text-based} & \multicolumn{4}{c}{STA-Subset ($n$=128) R@IoU} \\
 \multicolumn{1}{c}{LLM} & \multicolumn{1}{c|}{LLM} & \multicolumn{1}{c}{0.3} & \multicolumn{1}{c}{0.5} & \multicolumn{1}{c}{0.7} & \multicolumn{1}{c}{mIoU} \\
 \hline
 \multirow{4}{*}{LLaVA 7B} & \multicolumn{1}{l|}{GPT-4} & \centering \textbf{38.3} & \centering 14.8 & \centering 5.5 & \multicolumn{1}{c}{\textbf{26.3}} \\
 & \multicolumn{1}{l|}{Qwen 14B} & \centering 36.7 & \centering \textbf{17.2} & \centering 5.5 & \multicolumn{1}{c}{24.8} \\
 & \multicolumn{1}{l|}{Qwen 7B} & \centering 31.3 & \centering 10.9 & \centering 1.6 & \multicolumn{1}{c}{20.3} \\
 & \multicolumn{1}{l|}{Gemma 7B} & \centering 35.2 & \centering 11.7 & \centering 3.1 & \multicolumn{1}{c}{23.1} \\
 \hline
 \multirow{4}{*}{LLaVA 16B} & \multicolumn{1}{l|}{GPT-4} & \centering 28.1 & \centering 12.5 & \centering 4.7 & \multicolumn{1}{c}{21.3} \\
 & \multicolumn{1}{l|}{Qwen 14B} & \centering 24.2 & \centering 9.4 & \centering 1.6 & \multicolumn{1}{c}{18.2} \\
 & \multicolumn{1}{l|}{Qwen 7B} & \centering 28.1 & \centering 11.7 & \centering 3.9 & \multicolumn{1}{c}{19.2} \\
 & \multicolumn{1}{l|}{Gemma 7B} & \centering 29.7 & \centering 16.4 & \centering 3.1 & \multicolumn{1}{c}{21.0} \\
 \hline
 \multirow{4}{*}{GPT-4 Vision} & \multicolumn{1}{l|}{GPT-4} & \centering \textbf{44.5} & \centering \textbf{18.0} & \centering 7.0 & \multicolumn{1}{c}{\textbf{29.8}} \\
 & \multicolumn{1}{l|}{Qwen 14B} & \centering 33.6 & \centering 12.5 & \centering 4.7 & \multicolumn{1}{c}{22.9} \\
 & \multicolumn{1}{l|}{Qwen 7B} & \centering 40.6 & \centering 16.4 & \centering 5.5 & \multicolumn{1}{c}{25.1} \\
 & \multicolumn{1}{l|}{Gemma 7B} & \centering 28.9 & \centering 11.7 & \centering 3.9 & \multicolumn{1}{c}{19.1} \\
 \hline
 \multicolumn{2}{c|}{Gemini 1.5 Pro (Video)} & \centering 34.4 & \centering 16.4 & \centering \textbf{7.8} & \multicolumn{1}{c}{25.0} \\
 \hline\hline
 \multicolumn{2}{c|}{ViGA (VGG)} & \centering 56.2 & \centering 35.2 & \centering 14.8 & \multicolumn{1}{c}{35.8} \\
 \multicolumn{2}{c|}{ViGA (I3D)} & \centering 72.7 & \centering 45.3 & \centering 15.6 & \multicolumn{1}{c}{43.9} \\
 \hline
\end{tabular}
\vspace{.5em}
\captionsetup{justification=centering}
\caption{Comparing metrics (IoU) for different multimodal and text-based LLMs,\\
ViGA~\citep{cui2022video} and Gemini models using the \textit{activity prompting} strategy}
\label{table:perfmetrics}
\end{center}
\end{table}

\section{Analysis} \label{analysis}

\paragraph{Instruction-tuning and multimodal LLM outputs} We look at the qualitative effects of instruction-tuning a multimodal LLM to provide a more elaborate description of the scene. This will provide the text-based LLM with more information to better discern the interval in which the activity in question occurred. The following two text boxes show the description extracted from the scene in Figure \ref{activity_illustration} at timestep 00:19, where the individual is in the process of putting on their shoes.

\vspace{1em}

\begin{customtb}
\texttt{In the image, a person is standing in a doorway, holding a pair of pants in their hand. They are likely in the process of putting on or taking off the pants. The person is positioned near the center of the frame, and the doorway appears to be a part of a hallway.}
\end{customtb}
{\centering Base LLaVA 7B model output for Frame at 00:19 in Figure \ref{activity_illustration}\par}

\vspace{1em}

\begin{customtb}
\texttt{In the frame provided, we see a person in the process of putting on a pair of shoes. The individual is bending forward slightly, with their right hand reaching into a shoe, likely to place their foot inside. The shoe appears to be a dark-colored, possibly black, and the person is wearing a green t-shirt and purple pants. The setting looks like a hallway, with a closed door to the left and a light-colored wall in the background. The floor is a warm-colored wood, and the overall lighting suggests an indoor setting. The person's focus is on the task at hand, and their posture indicates they are engaged in the action of putting on their shoes.}
\end{customtb}
{\centering Instruction-tuned LLaVA 7B model output for Frame at 00:19 in Figure \ref{activity_illustration}\par}

\vspace{1em}

The example shows that the tuned model provides a more descriptive output regarding the scene. This includes detailed descriptions of the movements being conducted and the environment, all of which help in gathering context about when a particular activity starts or ends.

\paragraph{Text-based LLM outputs} In the text-based LLM setting, each model has its own quirks with regards to outputs. GPT-4 LLM models, while most accurate, simply generate a strict JSON file, making for easy parsing. Gemma 7B models are similar, providing a simple start and end timestamp without explanation. The Qwen 14B and 7B models are more verbose, and the JSON strings that are generated sometimes cannot be parsed using a standard json parser. If this is the case, heuristics are used to extract out time signatures embedded within the text, however, some time signatures may have been missed, negatively impacting performance.
\begin{customtb}
\texttt{Based on the descriptions provided, the action of "person putting on shoes" seems to have occurred in multiple instances, particularly from Frame 16 to Frame 28. However, the exact start and end frames are not explicitly stated. To provide a rough estimate, we can look for instances where the person is clearly seen bending over or interacting with their feet, which is a common action when putting on shoes. Here's a possible range for the start and end frames, considering the context and the actions described:\\
\\
\textasciigrave\textasciigrave\textasciigrave json\\
\{\\
  "start\_frame": "16",\\
  "end\_frame": "28"\\
\}\\
\textasciigrave\textasciigrave\textasciigrave
}
\end{customtb}

{\centering Temporal activity localization results from Qwen 7B\par}

\paragraph{Analyzing different input modes} In Figure~\ref{different_types}, we present prompting outputs for different types of visual input modes: a) using a single image, b) concatenating multiple consecutive images as a single image, and c) using a short 1-second video clip as input. We find that the multiple image and video modalities tend to give inconsistent descriptions, with the single image input mode providing the most detailed and accurate description. This may be due to the fact that multiple split images or 1-second clips are generally not as abundantly provided in the pretraining stage for these multimodal LLMs. Additionally, due to the fact that these models are relatively small (7B), it may even be a struggle to generate accurate descriptions for a single image, let alone multiple images or a video.

\section{Conclusion} \label{conclusion}

We demonstrate that instruction-tuning can be an effective way to generate elaborate and rich descriptions of scenes in smaller-sized multimodal LLMs. Our experiments show that a tuned multimodal LLM model can provide significant performance improvements in the task of temporal activity localization.

Our proposed two-stage approach, using only pretrained models, performs better than the state-of-the-art video-based LLMs, although it does not outperform state-of-the-art vision models specifically designed for temporal activity localization. The advantage over video-based LLMs is likely due to the higher accuracy and detail provided by image-based multimodal LLMs, as they are currently better trained and developed. Notably, the LLaVA 7B multimodal LLM performs well in the task of temporal grounding of activities, even exceeding the performance of Gemini 1.5 Pro when coupled with a strong text-based LLM.

While our current work focused on a single description of a scene, an obvious next step is to tap into the reasoning capabilities of the LLM. Our instruction-tuning dataset includes additional Q\&A regarding actions within the scene, which, through further instruction-tuning, is expected to help the model gain additional insight into the actions occurring in the scene. Furthermore, exploring methods such as chain of thought~\citep{scienceqa} or tree of thought~\citep{yao2024tree} prompting could be a valuable direction for tapping into the reasoning capabilities of the LLM and improving performance.

\pagebreak

\bibliographystyle{acl_natbib}
\bibliography{references}

\begin{thebibliography}{28}
\expandafter\ifx\csname natexlab\endcsname\relax\def\natexlab#1{#1}\fi

\bibitem[{Alayrac et~al.(2022)Alayrac, Donahue, Luc, Miech, Barr, Hasson, Lenc, Mensch, Millican, Reynolds et~al.}]{alayrac2022flamingo}
Jean-Baptiste Alayrac, Jeff Donahue, Pauline Luc, Antoine Miech, Iain Barr, Yana Hasson, Karel Lenc, Arthur Mensch, Katherine Millican, Malcolm Reynolds, et~al. 2022.
\newblock Flamingo: a visual language model for few-shot learning.
\newblock \emph{Advances in Neural Information Processing Systems}, 35:23716--23736.

\bibitem[{Anne~Hendricks et~al.(2017)Anne~Hendricks, Wang, Shechtman, Sivic, Darrell, and Russell}]{anne2017localizing}
Lisa Anne~Hendricks, Oliver Wang, Eli Shechtman, Josef Sivic, Trevor Darrell, and Bryan Russell. 2017.
\newblock Localizing moments in video with natural language.
\newblock In \emph{Proceedings of the IEEE international conference on computer vision}, pages 5803--5812.

\bibitem[{Bai et~al.(2023)Bai, Bai, Chu, Cui, Dang, Deng, Fan, Ge, Han, Huang, Hui, Ji, Li, Lin, Lin, Liu, Liu, Lu, Lu, Ma, Men, Ren, Ren, Tan, Tan, Tu, Wang, Wang, Wang, Wu, Xu, Xu, Yang, Yang, Yang, Yang, Yao, Yu, Yuan, Yuan, Zhang, Zhang, Zhang, Zhang, Zhou, Zhou, Zhou, and Zhu}]{qwen}
Jinze Bai, Shuai Bai, Yunfei Chu, Zeyu Cui, Kai Dang, Xiaodong Deng, Yang Fan, Wenbin Ge, Yu~Han, Fei Huang, Binyuan Hui, Luo Ji, Mei Li, Junyang Lin, Runji Lin, Dayiheng Liu, Gao Liu, Chengqiang Lu, Keming Lu, Jianxin Ma, Rui Men, Xingzhang Ren, Xuancheng Ren, Chuanqi Tan, Sinan Tan, Jianhong Tu, Peng Wang, Shijie Wang, Wei Wang, Shengguang Wu, Benfeng Xu, Jin Xu, An~Yang, Hao Yang, Jian Yang, Shusheng Yang, Yang Yao, Bowen Yu, Hongyi Yuan, Zheng Yuan, Jianwei Zhang, Xingxuan Zhang, Yichang Zhang, Zhenru Zhang, Chang Zhou, Jingren Zhou, Xiaohuan Zhou, and Tianhang Zhu. 2023.
\newblock Qwen technical report.
\newblock \emph{arXiv preprint arXiv:2309.16609}.

\bibitem[{Chen et~al.(2023{\natexlab{a}})Chen, Zhu, Shen, Li, Liu, Zhang, Krishnamoorthi, Chandra, Xiong, and Elhoseiny}]{chen2023minigptv2}
Jun Chen, Deyao Zhu, Xiaoqian Shen, Xiang Li, Zechun Liu, Pengchuan Zhang, Raghuraman Krishnamoorthi, Vikas Chandra, Yunyang Xiong, and Mohamed Elhoseiny. 2023{\natexlab{a}}.
\newblock \href {http://arxiv.org/abs/2310.09478} {Minigpt-v2: large language model as a unified interface for vision-language multi-task learning}.

\bibitem[{Chen et~al.(2023{\natexlab{b}})Chen, Spiridonova, Yang, Gao, and Li}]{chen2023llava}
Wei-Ge Chen, Irina Spiridonova, Jianwei Yang, Jianfeng Gao, and Chunyuan Li. 2023{\natexlab{b}}.
\newblock Llava-interactive: An all-in-one demo for image chat, segmentation, generation and editing.
\newblock \emph{arXiv preprint arXiv:2311.00571}.

\bibitem[{Cui et~al.(2022)Cui, Qian, Peng, Daskalaki, Chen, Guo, Sun, and Jiang}]{cui2022video}
Ran Cui, Tianwen Qian, Pai Peng, Elena Daskalaki, Jingjing Chen, Xiaowei Guo, Huyang Sun, and Yu-Gang Jiang. 2022.
\newblock Video moment retrieval from text queries via single frame annotation.
\newblock In \emph{Proceedings of the 45th International ACM SIGIR Conference on Research and Development in Information Retrieval}, pages 1033--1043.

\bibitem[{Gao et~al.(2017)Gao, Sun, Yang, and Nevatia}]{gao2017tall}
Jiyang Gao, Chen Sun, Zhenheng Yang, and Ram Nevatia. 2017.
\newblock Tall: Temporal activity localization via language query.
\newblock In \emph{Proceedings of the IEEE international conference on computer vision}, pages 5267--5275.

\bibitem[{Huang et~al.(2023)Huang, Wang, Chen, Song, and Zhu}]{huang2023vtimellm}
Bin Huang, Xin Wang, Hong Chen, Zihan Song, and Wenwu Zhu. 2023.
\newblock Vtimellm: Empower llm to grasp video moments.
\newblock \emph{arXiv preprint arXiv:2311.18445}.

\bibitem[{Li et~al.(2017)Li, Jabri, Joulin, and Van Der~Maaten}]{li2017learning}
Ang Li, Allan Jabri, Armand Joulin, and Laurens Van Der~Maaten. 2017.
\newblock Learning visual n-grams from web data.
\newblock In \emph{Proceedings of the IEEE International Conference on Computer Vision}, pages 4183--4192.

\bibitem[{Li et~al.(2023{\natexlab{a}})Li, Li, Savarese, and Hoi}]{li2023blip}
Junnan Li, Dongxu Li, Silvio Savarese, and Steven Hoi. 2023{\natexlab{a}}.
\newblock Blip-2: Bootstrapping language-image pre-training with frozen image encoders and large language models.
\newblock \emph{arXiv preprint arXiv:2301.12597}.

\bibitem[{Li et~al.(2023{\natexlab{b}})Li, He, Wang, Li, Wang, Luo, Wang, Wang, and Qiao}]{li2023videochat}
KunChang Li, Yinan He, Yi~Wang, Yizhuo Li, Wenhai Wang, Ping Luo, Yali Wang, Limin Wang, and Yu~Qiao. 2023{\natexlab{b}}.
\newblock Videochat: Chat-centric video understanding.
\newblock \emph{arXiv preprint arXiv:2305.06355}.

\bibitem[{Li et~al.(2024)Li, Xu, Zhang, Song, Cai, Qi, Zhou, Pan, Li, Vu, Huang, and Wang}]{li2024groundinggptlanguage}
Zhaowei Li, Qi~Xu, Dong Zhang, Hang Song, Yiqing Cai, Qi~Qi, Ran Zhou, Junting Pan, Zefeng Li, Van~Tu Vu, Zhida Huang, and Tao Wang. 2024.
\newblock \href {http://arxiv.org/abs/2401.06071} {Groundinggpt:language enhanced multi-modal grounding model}.

\bibitem[{Liu et~al.(2023{\natexlab{a}})Liu, Li, Li, and Lee}]{liu2023improved}
Haotian Liu, Chunyuan Li, Yuheng Li, and Yong~Jae Lee. 2023{\natexlab{a}}.
\newblock Improved baselines with visual instruction tuning.
\newblock \emph{arXiv preprint arXiv:2310.03744}.

\bibitem[{Liu et~al.(2023{\natexlab{b}})Liu, Li, Wu, and Lee}]{liu2023llava}
Haotian Liu, Chunyuan Li, Qingyang Wu, and Yong~Jae Lee. 2023{\natexlab{b}}.
\newblock Visual instruction tuning.
\newblock In \emph{NeurIPS}.

\bibitem[{Lu et~al.(2022)Lu, Mishra, Xia, Qiu, Chang, Zhu, Tafjord, Clark, and Kalyan}]{scienceqa}
P.~Lu, S.~Mishra, T.~Xia, L.~Qiu, K.W. Chang, S.C. Zhu, O.~Tafjord, P.~Clark, and A.~Kalyan. 2022.
\newblock Learn to explain: Multimodal reasoning via thought chains for science question answering.
\newblock In \emph{The 36th Conference on Neural Information Processing Systems (NeurIPS)}.

\bibitem[{Maaz et~al.(2023)Maaz, Rasheed, Khan, and Khan}]{maaz2023video}
Muhammad Maaz, Hanoona Rasheed, Salman Khan, and Fahad~Shahbaz Khan. 2023.
\newblock Video-chatgpt: Towards detailed video understanding via large vision and language models.
\newblock \emph{arXiv preprint arXiv:2306.05424}.

\bibitem[{OpenAI(2023)}]{openai2023gpt4}
OpenAI. 2023.
\newblock \href {http://arxiv.org/abs/2303.08774} {Gpt-4 technical report}.

\bibitem[{Qian et~al.(2024)Qian, Li, Wu, Ye, Fei, Chua, Zhuang, and Tang}]{qian2024momentor}
Long Qian, Juncheng Li, Yu~Wu, Yaobo Ye, Hao Fei, Tat-Seng Chua, Yueting Zhuang, and Siliang Tang. 2024.
\newblock \href {http://arxiv.org/abs/2402.11435} {Momentor: Advancing video large language model with fine-grained temporal reasoning}.

\bibitem[{Radford et~al.(2021)Radford, Kim, Hallacy, Ramesh, Goh, Agarwal, Sastry, Askell, Mishkin, Clark et~al.}]{radford2021learning}
Alec Radford, Jong~Wook Kim, Chris Hallacy, Aditya Ramesh, Gabriel Goh, Sandhini Agarwal, Girish Sastry, Amanda Askell, Pamela Mishkin, Jack Clark, et~al. 2021.
\newblock Learning transferable visual models from natural language supervision.
\newblock In \emph{International conference on machine learning}, pages 8748--8763. PMLR.

\bibitem[{Reid et~al.(2024)Reid, Savinov, Teplyashin, Lepikhin, Lillicrap, Alayrac, Soricut, Lazaridou, Firat, Schrittwieser et~al.}]{reid2024gemini}
Machel Reid, Nikolay Savinov, Denis Teplyashin, Dmitry Lepikhin, Timothy Lillicrap, Jean-baptiste Alayrac, Radu Soricut, Angeliki Lazaridou, Orhan Firat, Julian Schrittwieser, et~al. 2024.
\newblock Gemini 1.5: Unlocking multimodal understanding across millions of tokens of context.
\newblock \emph{arXiv preprint arXiv:2403.05530}.

\bibitem[{Sigurdsson et~al.(2016)Sigurdsson, Varol, Wang, Laptev, Farhadi, and Gupta}]{sigurdsson2016hollywood}
Gunnar~A. Sigurdsson, G{\"u}l Varol, Xiaolong Wang, Ivan Laptev, Ali Farhadi, and Abhinav Gupta. 2016.
\newblock \href {http://arxiv.org/abs/1604.01753} {Hollywood in homes: Crowdsourcing data collection for activity understanding}.
\newblock \emph{ArXiv e-prints}.

\bibitem[{Touvron et~al.(2023)Touvron, Lavril, Izacard, Martinet, Lachaux, Lacroix, Rozi{\`e}re, Goyal, Hambro, Azhar, Rodriguez, Joulin, Grave, and Lample}]{touvron2023llama}
H.~Touvron, T.~Lavril, G.~Izacard, X.~Martinet, M.A. Lachaux, T.~Lacroix, B.~Rozi{\`e}re, N.~Goyal, E.~Hambro, F.~Azhar, A.~Rodriguez, A.~Joulin, E.~Grave, and G.~Lample. 2023.
\newblock Llama: Open and efficient foundation language models.
\newblock \url{https://arxiv.org/abs/2302.13971}.
\newblock ArXiv preprint arXiv:2302.13971.

\bibitem[{Wang et~al.(2020)Wang, Ma, and Jiang}]{wang2020temporally}
Jingwen Wang, Lin Ma, and Wenhao Jiang. 2020.
\newblock Temporally grounding language queries in videos by contextual boundary-aware prediction.
\newblock In \emph{Proceedings of the AAAI Conference on Artificial Intelligence}, volume~34, pages 12168--12175.

\bibitem[{Yao et~al.(2024)Yao, Yu, Zhao, Shafran, Griffiths, Cao, and Narasimhan}]{yao2024tree}
Shunyu Yao, Dian Yu, Jeffrey Zhao, Izhak Shafran, Tom Griffiths, Yuan Cao, and Karthik Narasimhan. 2024.
\newblock Tree of thoughts: Deliberate problem solving with large language models.
\newblock \emph{Advances in Neural Information Processing Systems}, 36.

\bibitem[{Yuan et~al.(2019)Yuan, Ma, Wang, Liu, and Zhu}]{yuan2019semantic}
Yitian Yuan, Lin Ma, Jingwen Wang, Wei Liu, and Wenwu Zhu. 2019.
\newblock Semantic conditioned dynamic modulation for temporal sentence grounding in videos.
\newblock \emph{Advances in Neural Information Processing Systems}, 32.

\bibitem[{Zhang et~al.(2023)Zhang, Li, and Bing}]{zhang2023video}
Hang Zhang, Xin Li, and Lidong Bing. 2023.
\newblock Video-llama: An instruction-tuned audio-visual language model for video understanding.
\newblock \emph{arXiv preprint arXiv:2306.02858}.

\bibitem[{Zhang et~al.(2024)Zhang, Han, Liu, Zhou, Lu, Li, Gao, and Qiao}]{zhang2024llamaadapter}
Renrui Zhang, Jiaming Han, Chris Liu, Aojun Zhou, Pan Lu, Hongsheng Li, Peng Gao, and Yu~Qiao. 2024.
\newblock \href {https://openreview.net/forum?id=d4UiXAHN2W} {{LL}a{MA}-adapter: Efficient fine-tuning of large language models with zero-initialized attention}.
\newblock In \emph{The Twelfth International Conference on Learning Representations}.

\bibitem[{Zhang et~al.(2020)Zhang, Peng, Fu, and Luo}]{zhang2020learning}
Songyang Zhang, Houwen Peng, Jianlong Fu, and Jiebo Luo. 2020.
\newblock Learning 2d temporal adjacent networks for moment localization with natural language.
\newblock In \emph{Proceedings of the AAAI Conference on Artificial Intelligence}, volume~34, pages 12870--12877.

\end{thebibliography}

\end{document}